\documentclass[usletter, 10pt, conference]{ieeeconf}

\usepackage{url}

\usepackage[ruled,noend,linesnumbered]{algorithm2e}
\usepackage{graphicx,color,psfrag}
\usepackage{amsmath}
\usepackage{amsfonts}
\usepackage{amssymb}
\usepackage{bigdelim}
\usepackage{dsfont}
\usepackage{color, soul}
\usepackage{graphicx}
\usepackage{caption}
\usepackage{subcaption}

\IEEEoverridecommandlockouts              
\title{High-quality Instance-aware Semantic 3D Map Using RGB-D Camera} 

\author{Dinh-Cuong Hoang$^{*}$, Todor Stoyanov$^{*}$, and Achim J. Lilienthal$^{*}$
\thanks{*Centre for Applied Autonomous Sensor Systems (AASS); Orebro University.}
}

\begin{document}
\maketitle
\thispagestyle{empty}
\pagestyle{empty}


\begin{abstract}

We present a mapping system capable of constructing detailed instance-level semantic models of room-sized indoor environments by means of an RGB-D camera. In this work, we integrate deep-learning based instance segmentation and classification into a state of the art RGB-D SLAM system. We leverage the pipeline of ElasticFusion \cite{whelan2016elasticfusion} as a backbone, and propose modifications of the registration cost function. The proposed objective function features a tunable weight for the appearance channel, which can be learned from data. The resulting system is capable of producing accurate semantic maps of room-sized environments, as well as reconstructing highly detailed object-level models. The developed method has been verified through experimental validation on the TUM RGB-D SLAM benchmark and the YCB video dataset. Our results confirmed that the proposed system performs favorably in terms of trajectory estimation, surface reconstruction, and segmentation quality in comparison to other state-of-the-art systems.

\end{abstract}


\section{Introduction}
\label{sec:intro}

With the advent of 3D measurement devices using structured light sensing such as affordable RGB-D cameras like the ASUS Xtion Pro Live or Microsoft's Kinect, research on SLAM (Simultaneous Localization and Mapping) has made giant strides in development \cite{whelan2016elasticfusion, newcombe2011kinectfusion, canelhas2013sdf}. These approaches achieve dense surface reconstruction of complex indoor scenes while maintaining real-time performance through implementations on highly parallelized hardware. Beyond classical SLAM systems which solely provide a purely geometric map, the idea of a system that generates a dense map in which object instances are semantically annotated has attracted substantial interest in the research community \cite{nakajima2019efficient, sunderhauf2017meaningful, mccormac2017semanticfusion, mccormac2018fusion++, runz2018maskfusion}. An instance-aware semantic 3D map is useful for enabling more context-aware and more intelligent robot behaviors.
\begin{figure}
\centering
	\includegraphics[width = 0.95\linewidth, height=4.5cm]{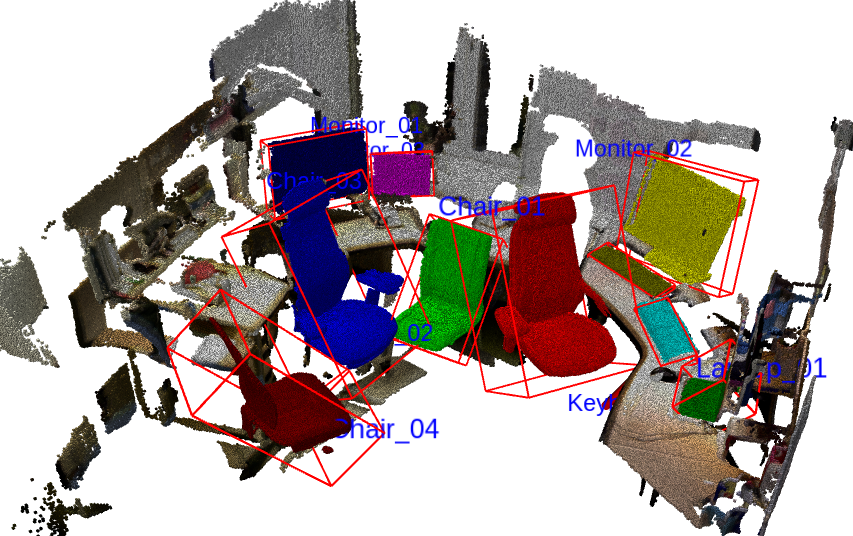}
\caption{An instance-aware semantic 3D map of our office produced by the proposed mapping system.}
\label{fig:overview}
\end{figure}
%
\par
In this study, we propose a 3D mapping system to produce highly accurate object-aware semantic scene reconstruction. Our work benefits from incorporating state of the art RGB-D SLAM and deep-learning-based instance segmentation techniques \cite{whelan2016elasticfusion, he2017mask}. We develop a CNN architecture beyond the original Mask R-CNN to input RGB image and output an adaptive weight for the cost function used in the sensor pose estimation process. In contrast to existing approaches that update the probabilities for all elements (surfels or voxels) in the 3D map, we reduce the space complexity by a more efficient strategy based on instance labels. In addition to the highly accurate semantic scene reconstruction, we correct misclassified regions using two proposed criteria which rely on location information and pixel-wise probability of the class. We evaluate the performance of our system on the TUM RGB-SLAM benchmark and the YCB video dataset \cite{xiang2017posecnn} and show that our system benefits greatly from the use of the proposed joint cost function with adaptive weights. The developed system performs on par with the state of the art in terms of camera trajectory estimation while generating accurate object instance models. We also show that our approach leads to an improvement in the 2D instance labeling over baseline single frame predictions. 
\section{RELATED WORK}
\subsection{Registration of RGB-D Images}

A large number of registration algorithms have been proposed in the context of RGB-D Tracking and Mapping (TAM) \cite{whelan2016elasticfusion, newcombe2011kinectfusion, canelhas2013sdf, huang2017visual, endres2012evaluation}. Feature-based approaches estimate the sensor pose by only considering informative and characteristic points known as key points \cite{huang2017visual, endres2012evaluation}. 
	Alternatively, dense geometric tracking approaches, such as KinectFusion \cite{newcombe2011kinectfusion}, typically apply an ICP \cite{chen1992object} variant to directly register the full depth image to an online reconstructed volumetric model.
	The original KinectFusion algorithm uses a Truncated Signed Distance Function (TSDF) \cite{curless1996volumetric} for model representation and point-to-plane ICP \cite{chen1992object} for alignment. Several alternatives to this choice of algorithms have been proposed \cite{whelan2012kintinuous, whelan2015real, munoz2016point}, which are expected to perform better in regions where the point-to-plane distance is ill-defined. 
\par
Using only depth data, tracking failure can occur in situations where the amount of characteristic features in the depth map is low. Steinbrucker et al. \cite{steinbrucker2011real} introduced an energy minimization approach for RGB-D image registration that relies on color information instead. In comparison with geometric ICP, the authors reported that their method is more accurate in the regime of small camera motions.
	Whelan et al. \cite{whelan2013robust} combined the color and depth information in the cost function so that all given information is used. They demonstrated that this combination increases the robustness of camera tracking across a variety of environments. This idea was further used in ElasticFusion \cite{whelan2016elasticfusion} which fuses measurements and uses a surfel structure instead of volumetric one for reconstruction. ElasticFusion demonstrates the capability to produce globally consistent reconstructions in real-time without the use of post-processing steps. Similarly to Elastic Fusion,  our approach also integrates both geometric and photometric cues for camera tracking. 

\subsection{Semantic Mapping}

Fusing semantic along with geometry information within a 3D reconstructed map is a promising approach to enable intelligent systems to better understand a 3D scene. A number of semantic mapping systems have been developed \cite{mccormac2017semanticfusion, hermans2014dense, wong2017segicp}.
	Hermans et al. \cite{hermans2014dense} utilize Random Decision Forests to achieve semantic pixel-wise image labeling and fuse them in a classic Bayesian framework. 
	Previous work by McCormac et al. \cite{mccormac2017semanticfusion} aimed towards a useful semantic 3D map by combining the advantages of Convolutional Neural Networks (CNNs) and ElasticFusion \cite{whelan2016elasticfusion}. The correspondences between frames are estimated by the SLAM system. Meanwhile, their CNN architecture adopts a Deconvolutional Semantic Segmentation network \cite{noh2015learning} to generate a pixel-wise semantic map for incoming images. 
	Unlike the original architecture \cite{noh2015learning}, this system incorporates depth information to obtain a higher accuracy than the pretrained RGB network. The authors reported that fusing multiple predictions led to a significant improvement in the semantic labeling and it is the first real-time capable approach suitable for interactive indoor scene scanning and labeling. 
	Likewise, SegICP-DSR \cite{wong2017segicp} fuses RGB-D observations into a semantically-labeled point cloud for object pose estimation using adversarial networks and ElasticFusion. There is, however, one significant difference. SegICP-DSR employs the semantic label difference instead of a photometric error when formulating the alignment objective function. Then, a semantically-labeled point cloud can be directly outputted from the reconstruction process without an extra update step. 
	Obviously, the addition of semantic information enables a much greater range of functionality than geometry alone. However, since the above systems only consider class labels, they are limited to scenarios with single object instances per scene and may degenerate performance in case multiple objects of the same type are present.
\par
MaskFusion \cite{runz2018maskfusion} is a real-time, object-aware, semantic and dynamic RGB-D SLAM system. It combines geometric segmentation running on every frame and semantic segmentation using Mask R-CNN computed for select keyframes. The geometric segmentation algorithm acquires object boundaries based on an analysis of depth discontinuities and surface normals, while Mask R-CNN is used to provide object masks with semantic labels. Camera poses are estimated by minimizing a joint geometric and photometric error function as presented in \cite{whelan2016elasticfusion}. The reported results demonstrate that while MaskFusion outperforms a set of baseline state of the art algorithms in highly dynamic scenes, ElasticFusion performs best on static and moderately dynamic scenes. Unlike MaskFusion, our system is not designed to deal with dynamic scenes and real-time operation. Instead, we assume all objects to be static during an observation and aim to reconstruct high-quality object models.
\par
Most related to ours is the work of McCormac et al. Fusion++ \cite{mccormac2018fusion++}, which aims to produce multiple semantically labeled maps of object instances without a dense representation of the entire static scene. Fusion++ uses Mask R-CNN instance segmentation to initialize dense per-object TSDF reconstructions with object size-dependent resolutions. For camera tracking, Fusion++ takes an approach similar to KinectFusion using projective data association and a point-to-plane error. Note that apart from object level maps, Fusion++ also maintains a coarse background TSDF to assist frame-to-model tracking. While the authors evaluated the trajectory error of the developed system against the baseline approach of simple coarse TSDF odometry, the reports did not provide a comparison with other photometry or semantics-aware state of the art approaches.
\par
Though we aim towards multiple object level maps, there are clear differences between Fusion++ and our work. While Fusion++ only uses depth information for motion tracking, the proposed system increases the robustness of sensor tracking through integrating appearance and geometric cues into the reconstruction process. In addition, our CNN network is able to generate an adaptive weight for the joint cost function which plays a big role in robust camera pose tracking.
%

\section{METHODOLOGY}
\label{sec:methodology}

\begin{figure}
\centering
	\includegraphics[width = 0.95\linewidth, height=4.5cm]{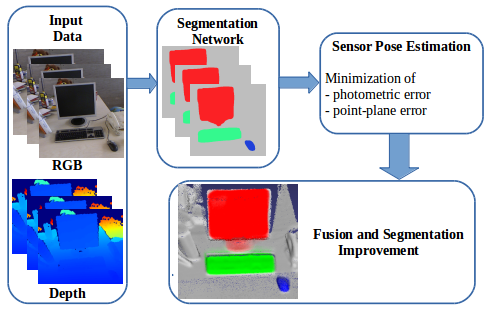}
\caption{\textbf{Flow of the proposed framework}: The segmentation network firstly yields masks and probabilities specified for each category. Then the output of the segmentation stage along with depth map and RGB frame is used for camera pose estimation. Finally, input data and semantic information are fused into the 3D map based on the transformation matrix estimated from the previous stage.}
\label{fig:overview}
\end{figure}

\begin{figure*}[t!]
\centering
	\includegraphics[width = 0.95\linewidth, height=5.5cm]{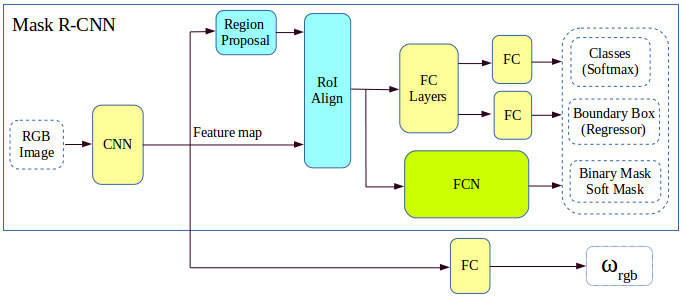}
\caption{\textbf{CNN architecture}: Extending Mask R-CNN to predict masks and classes probabilities while simultaneously yielding an adaptive weight for camera tracking.}
\label{fig:CNN-Architecture}
\end{figure*}

Our pipeline is composed of three main components as illustrated in Fig.~\ref{fig:overview}. The input RGB-D data is processed through a semantic instance segmentation stage, followed by a frame-to-model alignment stage, and finally a model fusion stage. In the following, we summarise the key elements of our method.

\textbf{\textit{Segmentation}:} Produce object masks with semantic labels using our CNN architecture. The developed architecture also predicts weights for the joint cost function for camera tracking. 

\textbf{\textit{Tracking}:} Estimate camera poses within the ElasticFusion pipeline using our proposed joint cost function. We combine the cost functions of geometric and photometric estimates in a weighted sum. The adaptive weight for color image is generated by the segmentation process above. 

\textbf{\textit{Fusion and Segmentation Improvement}:} Fuse segmentation information into 3D map using our instance-based semantic fusion. To improve segmentation accuracy, misclassified regions are corrected by two criteria which rely on a sequence of CNN predictions.

\subsection{Segmentation Network}
\label{sec:seg_net}

In our framework, we employ an end-to-end CNN framework, Mask R-CNN for generating a high-quality segmentation mask for each instance. Mask R-CNN has three outputs for each candidate object, a class label, a bounding-box offset, and a mask. Its procedure consists of two stages. In the first stage, candidate object bounding boxes are proposed by a Region Proposal Network (RPN). In the second stage, classification, bounding-box regression, and mask prediction are performed in parallel on each small feature map, which is extracted by RoIPool. Note that to speed up inference and improve accuracy the mask branch is applied to the highest scoring 100 detection boxes after running the box prediction. The mask branch predicts a binary mask from each RoI using an FCN architecture \cite{long2015fully}. The binary mask is a single $ m \times m $ output regardless of class, which is generated by binarizing the floating-number mask or soft mask at a threshold of 0.5.
\par
To integrate deep-learning based segmentation and classification into our system, we extend Mask R-CNN to identify object outlines at the pixel level while simultaneously generating an adaptive weight used in camera pose tracking stage as shown in Fig.~\ref{fig:CNN-Architecture}. A fourth branch is added to our CNN framework, which shares computation of feature maps with Mask R-CNN branches and outputs the weight by a fully connected layer. In general, the network consists of a backbone CNN, a region proposal network (RPN), an ROI classifier, a bounding Box Regressor, a mask branch, and a camera tracking weight estimator. The CNN backbone is a standard convolutional neural network that is used for extracting a feature map. This convolutional feature map not only becomes the input for the other stages of Mask R-CNN but also shares computation with our extended branch for adaptive weight estimation. Therefore, the developed network receives an RGB image and returns a set of per-pixel class probabilities and weights used in the cost function in the subsequent alignment stage. The weight estimation is treated as a classification problem where the target is a binary decision whether or not the given RGB image should be used in the registration process. In other words, we aim to train our weight predicting model as a binary classifier, where one class signifies that the RGB image contains useful information for the subsequent registration process, while the other class indicates the converse. The probability predicted from the classification model is considered as an adaptive weight for our joint cost function for camera pose estimation.

\subsection{Camera Tracking}
\label{sec:cam_track}

To perform camera tracking, our object-oriented mapping system maintains a fused surfel-based model of the environment (similar to the model used by ElasticFusion \cite{whelan2016elasticfusion}). Here we borrow and extend the notation proposed in the original ElasticFusion paper. The model is represented by a cloud of surfels $\mathcal{M}^{s}$, where each surfel consists of of a position
$p \in \mathbb{R}^{3}$ , normal $n \in \mathbb{R}^{3}$, colour $c \in \mathbb{N}^{3}$, weight $w \in \mathbb{R}$, radius $r \in \mathbb{R}$, initialisation timestamp $t_{0}$ and last updated timestamp $t$. In addition, each surfel aslo stores a object instance label $l_{\mathbf{o}} \in N$. Each object instance $\mathbf{o}$ is associated with a discrete probability distribution over potential class labels, $P(l_{\mathbf{o}}=l_{i})$ over the set of class labels, $l_{i}\in\mathcal{L}$.
\par
The image space domain is defined as $\Omega\subset{\mathbb{N}^{2}}$ , where an RGB-D frame is composed of a color map and a depth map $D$ of depth pixels $d: \Omega \rightarrow R$. We define the 3D back projection of a point $ u\in\Omega $ given a depth map $D$ as $p(u, D)=K^{-1}\textit{u}d(u)$ , where $K$ is the camera intrinsics matrix and $\textit{u}$ the homogeneous form of $u$. The perspective projection of a 3D point  $p=[x, y, z]^{\top}$ is defined as $u=\pi(Kp)$, where $\pi(p)=(x/z, y/z)$. In the following, we describe our proposed approach for combined ICP pose estimation.
\par
Our approach aims to estimate a sensor pose that minimizes the cost over a combination of the global point-plane energy and photometric error. We wish to minimize a joint optimization objective:
\begin{align}
E_{combined} = E_{icp} + \omega_{rgb}E_{rgb} \
\label{eq:cost_function}
\end{align}

\noindent where $E_{icp}$ and $\omega_{rgb}E_{rgb}$ are the geometric and photometric terms respectively. The photometric error function is weighted by factor predicted from our CNN.

Finally, we find the transformation by minimizing the objective (1) through the Gauss-Newton non-linear least-square method with a three-level coarse-to-fine pyramid scheme.

\subsection{Fusion and Segmentation Improvement}
\label{sec:seg_refine}

\textbf{Class labels fusion:} Given an RGB-D frame at time step $t$, each mask $M$ from Mask R-CNN must be corresponded to an instance in the 3D map. Otherwise, it will be assigned as a new instance. To find the corresponding instance, we use the tracked camera pose and existing instances in the map built at time step $t-1$ to predict binary masks via splatted rendering. The percent overlap between the mask $M$ and a predicted mask $\hat{M}$ for object instance $\mathbf{o}$ is computed as $\mathbb{U}(M, \hat{M})=\dfrac{M\cap\hat{M}}{\hat{M}}$. Then the mask $M$ is mapped to object instance $\mathbf{o}$ which has the predicted mask $\hat{M}$ with largest overlap, where $\mathbb{U}(M, \hat{M}) > 0.3$.

Unlike existing works \cite{mccormac2017semanticfusion, mccormac2018fusion++, runz2018maskfusion} that each element constituting 3D map such as surfel or TSDF stores a probability distribution over all classes, we propose to assign an object instance label $\mathbf{o}$ to each surfel and then this label is associated with a discrete probability distribution over potential class labels, $P(L_{\mathbf{o}}=l_{i})$ over the set of class labels, $l_{i}\in\mathbb{L}$. In consequence, we need only one probability vector for all surfels belonging to the same object entity. This makes a big difference when the number of surfels is much larger than the number of classes. To update the class probability distribution, means of a recursive Bayesian update is used in \cite{hermans2014dense}. However, this scheme often results in an overly confident class probability distribution that contains scores unsuitable for ranking in object detection \cite{mccormac2018fusion++}. In order to make the distribution become more even, we update the class probability by simple averaging:

\begin{align}
\ P(l_{i}|I_{1,..,t}) = \dfrac{1}{t} \sum_{j=1}^{t}(p_{j}|I_{t}) \
\label{eq:fusion_eq}
\end{align}

Moreover, previous related works miss the background/object probability from the binary mask branch that predicts which pixels correspond to the main classes (non-background), and which pixels correspond to the background. Conversely, we enrich segmentation information on each surfel by adding the probability to account for background/object predictions. To that end, each surfel in our 3D map has a non-background probability attribute $ p_{o} $.

As presented in \cite{he2017mask} the binary mask branch first generates a $m \times m $ floating-number mask which is then resized to the RoI size, and binarized at a threshold of 0.5. Therefore, we are able to extract a per-pixel non-background probability map with the same image size $480 \times 640$. Given the RGB-D frame at time step $t$, a non-background probability $ p_{\mathbf{o}}(I_{t}) $ is assigned to each pixel. Camera tracking and the 3D back projection introduced in section \ref{sec:cam_track} enables us to update all the surfels with the corresponding probability as following:

\begin{align}
\ p_{\mathbf{o}} = \frac{1}{t} \sum_{j=1}^{t}p_{j}(I_{t})\
\label{eq:fusion_eq}
\end{align}

\textbf{Segmentation Improvement:} Despite the power and flexibility of Mask R-CNN, it usually misclassified object boundary regions as background. In other words, the detailed structures of an object are often lost or smoothed. Thus, there is still much room for improvement in segmentation. We observe that many of the pixels in the misclassified regions have non-background probability just slightly smaller than 0.5, while the soft probabilities mask for real background pixel is often far below the threshold. Based on this observation, we expect to achieve a more accurate object-aware semantic scene reconstruction by considering non-background probability of surfels within a $n$ frames sequence. With this goal, each possible surfel $s$ ($0.4 < p_{\mathbf{o}} < 0.5$) is associated with a confidence $\vartheta(s)$. If a surfel is identified for the first time, its associated confidence is initialized to zero. Then, when a new frame arrives, we increment the confidence $\vartheta(s) \leftarrow \vartheta(s)+1$ only if the corresponding pixel of that surfel satisfies 2 criteria: (i) its non-background probability is greater than 0.4; (ii) there is at least one object pixel inside its 6-neighborhood. After $n$ frames, if the confidence $\vartheta(s)$ exceeds the threshold $\sigma_{object}$, we assign surfel $s$ to the closest instance. Otherwise, $\vartheta(s)$ is reset to zero. Here, we found $n=10$ and $\sigma_{object}=10$ provide good performance.

\section{EXPERIMENTS}
\label{sec:experiments}

We have evaluated our system by performing experiments on the TUM \cite{sturm2012benchmark} and YCB video \cite{xiang2017posecnn} datasets. These experiments are aimed at evaluating both trajectory estimation and surface reconstruction accuracy. A comparison against most related works is also performed here.

For all tests, we run our system on a standard desktop PC running 64-bit Ubuntu 16.04 Linux with an Intel Core i7-4770K 3.5GHz and a nVidia GeForce GTX 1080 Ti 6GB GPU. Our pipeline is implemented in Python with Tensorflow 1.6 for segmentation and C++ with CUDA for mapping. The input is standard $640 \times 480$ resolution RGB-D video.

To train our CNNs, We start with a weights file that’s been trained on the ImageNet dataset \cite{deng2009imagenet} with a ResNet-101 \cite{he2016deep} backbone. We finetune layers of Mask R-CNN on COCO dataset with 10 common object classes in indoor environments (backpack, chair, keyboard, laptop, monitor, computer mouse, cell phone,
sink, refrigerator, microwave) and on a portion of the YCB video data set not used in the evaluations. To train the weight estimator branch, we split SceneNN dataset \cite{hua2016scenenn} into two groups based on camera pose ground truth and trajectory estimation of ElasticFusion using only photometric error. 

\subsection{Camera Pose Tracking}
\label{sec:camera_pose_track}

We compare the trajectory estimation performance of our system to two most related works MaskFusion and Fusion++ on the widely used RGB-D benchmark of \cite{sturm2012benchmark}. This benchmark is one of the most popular datasets for the evaluation of RGB-D SLAM systems. The dataset covers a large variety of scenes and camera motions and provides sequences for debugging with slow motions as well as longer trajectories with and without loop closures. Each sequence contains the color and depth images, as well as the ground-truth trajectory from the motion capture system.  To evaluate the error in the estimated trajectory by comparing it with the ground-truth, we adopt the absolute trajectory error (ATE) root-mean-square error metric (RMSE) as proposed in \cite{sturm2012benchmark}.

\begin{table*}[h]
\caption{Comparison of ATE RMSE on RGB-D SLAM benchmark. All units given are in metres.}
\label{tab:ate_rmse}
\begin{center}
\begin{tabular}{|c|c|c|c|c|c|c|}
\hline
& PCL-KinFu & Kintinuous & ElasticFusion & MaskFusion & Fusion++ & Our System \\
\hline
freiburg1\_desk & 0.073 & 0.037 & \textbf{0.020} 
& 0.034 & 0.049 & 0.022 \\
freiburg1\_room & 0.187 & 0.075 & 0.068
& 0.153 & 0.235 & \textbf{0.065} \\
freiburg1\_desk2 & 0.102 & 0.071 & \textbf{0.048}
& 0.093 & 0.153 & 0.056 \\
freiburg1\_360 & - & 0.116 & \textbf{0.108}
& 0.157 & - & 0.126 \\
freiburg1\_teddy & - & 0.132 & \textbf{0.083}
& 0.129 & - & 0.095 \\
freiburg2\_desk & 0.103 & \textbf{0.034} & 0.071
& 0.108 & 0.114 & 0.083 \\
freiburg2\_xyz & 0.077 & 0.029 & \textbf{0.011}
& 0.041 & 0.020 & 0.025 \\
freiburg2\_rpy & - & 0.018 & 0.015
& 0.076 & - & \textbf{0.012} \\
freiburg3\_long\_office\_household & 0.086 & 0.030 & 0.017 
& 0.102 & 0.108 & 0.085 \\
freiburg3\_large\_cabinet & - & 0.144 & 0.099
& 0.133 & - & \textbf{0.052} \\
\hline
\end{tabular}
\end{center}
\end{table*}

Table~\ref{tab:ate_rmse} shows the results. From these we can see the performance of our system is comparable to state of the art classical approaches, and outperforms both MaskFusion and Fusion++. Results for Fusion++ are taken from the respective publication as presented by the authors, and values for MaskFusion are calculated from MaskFusion implementation. While the original ElasticFusion algorithm still obtains the best overall ATE performance, the results of our approach are comparable. Despite this relative similarity in the average trajectories, our approach performs better in reconstructing the relevant object-scale detail, as discussed in the next sub-section.

\subsection{Reconstruction}
\label{sec:reconstruc_result}

\begin{table}[h]
\caption{Comparison of surface reconstruction accuracy results on the YCB objects (mm).}
\label{tab:reconstruction}
\begin{center}
\begin{tabular}{|c|c|c|c|}
\hline
& ElasticFusion & MaskFusion & Our System \\
\hline
YCB video 0007 & 9.6 & 7.3 & \textbf{6.5} \\
YCB video 0036 & 8.1 & 6.4 & \textbf{5.7} \\
YCB video 0072 & 10.1 & 9.4 & \textbf{8.7} \\
Our sequence 01 & 7.1 & 6.7 & \textbf{3.7} \\
Our sequence 02 & 7.3 & 6.6 & \textbf{4.1} \\
Our sequence 03 & 7.5 & 6.2 & \textbf{3.4} \\
\hline
\end{tabular}
\end{center}
\end{table}
It should be noted that a good performance on a camera trajectory benchmark does not always imply a high quality surface reconstruction. We have evaluated our system by performing experiments on Yale-CMU-Berkeley (YCB) Object and Model set \cite{calli2015benchmarking}. We finetuned our network on the training set of the YCB-Video dataset. It contains 92 real video sequences for 21 object instances. 89 videos along with 80,000 synthetic images are used for training. We evaluate
on the remaining test videos from the original data set, as well as on three video sequences we acquired independently in scenes featuring a larger number of objects and more complex camera trajectories.

In order to evaluate surface reconstruction quality, we compare the object models obtained through our approach to the ground truth YCB object models. Note that the ground truth object models are only used here to compute evaluation metrics, unlike in prior works like SLAM++ which use them within the tracking framework. For every object present in the scene, we first register the reconstructed model M to the ground truth model G. Next, we project every vertex from M onto G, and compute the distance between the original vertex and it's projection. Finally, we calculate and report the mean distance $\mu_d$ over all model points and all objects.

Table~\ref{tab:reconstruction} shows the mean reconstruction error over the six sequences produced by our system, MaskFusion and ElasticFusion. Our method consistently results in the lowest reconstruction errors over all datasets. From this comparison it is evident that our approach benefits greatly from the use of the proposed joint cost function with adaptive weight. We also note that in our approach all surfels on objects of interest are always \textit{active}, while ElasticFusion segments these surfels into \textit{inactive} area if they have not been observed for a period of time $\partial_{t}$. This means that object surfels are updated all the time. As a result, our framework is able to produce a highly accurate object-oriented semantic map. 

\subsection{Segmentation Accuracy Evaluation}

\begin{figure}[t!]
\centering
	\includegraphics[width = 0.8\linewidth, height=4.5cm]{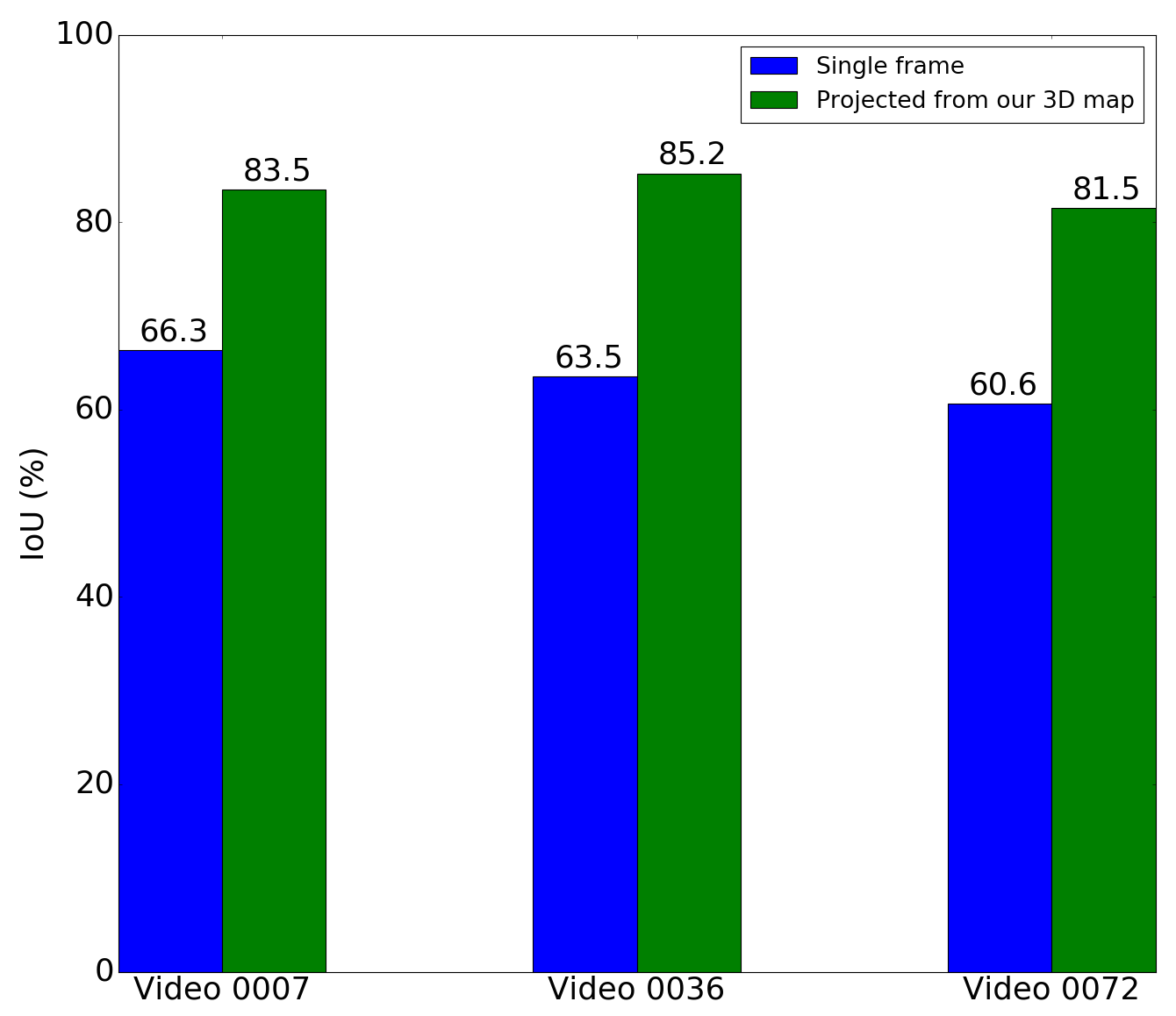}
\caption{Results of segmentation accuracy evaluation on YCB videos.}
\label{fig:seg-improve}
\end{figure}

In this section, we show on the YCB video dataset that our system leads to an improvement in the 2D instance labeling over the baseline single frame predictions generated by Mask R-CNN. Our 2D masks are obtained by reprojecting the reconstructed 3D model. We use the Intersection over Union (IoU) metric for this evaluation, which measures the number of pixels common between the ground-truth and prediction masks divided by the total number of pixels present across both masks. The results of this evaluation are shown in Fig.~\ref{fig:seg-improve}. We observe the segmentation performance improved, on average, from 63.5\% for a single frame to 83.4\% when projecting the predictions from the 3D map.

\subsection{Run-time Performance and Memory Usage}

\begin{figure}[t!]
\centering
	\includegraphics[width = 0.8\linewidth, height=4.5cm]{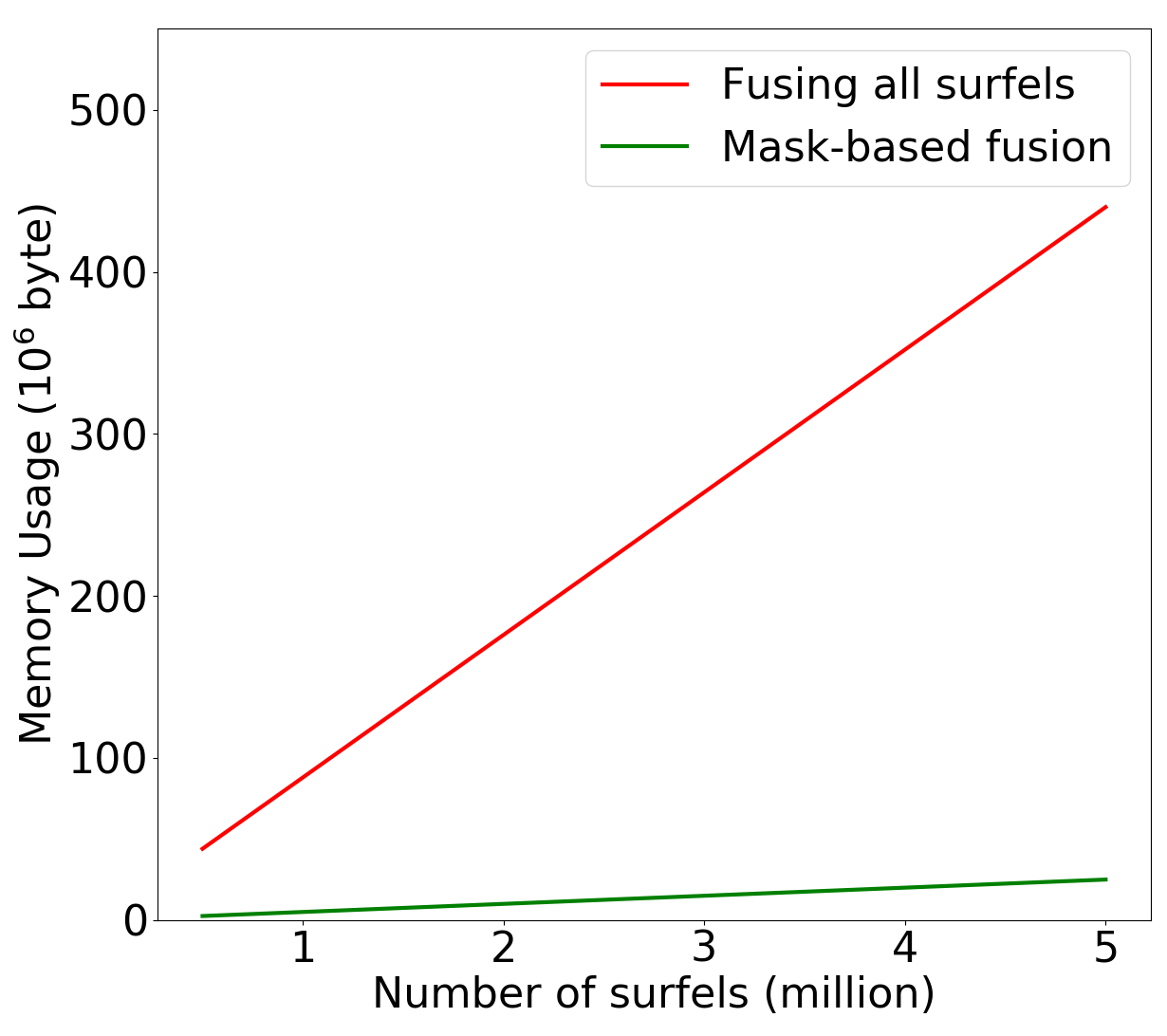}
\caption{Memory usage for storing class probabilities.}
\label{fig:memory_usage}
\end{figure}

\textbf{Run-time Performance:} Our current system does not run in real time because of heavy computation in instance segmentation. Our CNN requires 350ms, while camera pose estimation, fusion, and segmentation require a further 70ms on a typical sample of RGB-D SLAM Benchmark \cite{sturm2012benchmark}.

\textbf{Memory Usage:} We compared our mask-based fusion method with other approaches \cite{mccormac2017semanticfusion, mccormac2018fusion++, runz2018maskfusion} which assign class probabilities to each element of the 3D map rather than to each mask. The memory usage of the proposed method is significantly reduced compared to the conventional approach over all samples as shown in Fig.~\ref{fig:memory_usage}. The average memory usage of the proposed method is 5.7\% of those conventional approaches.

\section{CONCLUSIONS}

In this paper we have presented a 3D mapping system for RGB-D camera pose tracking that yields high quality object-oriented semantic reconstruction. Our system is based on incorporating state of the art RGB-D SLAM and deep-learning-based instance segmentation and classification. The developed system shows that by combining appearance and semantic cues with the adaptive weight in camera pose tracking we are able to obtain reliable camera tracking and state of the art surface reconstruction in small-scale environments populated with objects of interest. In addition, we propose an approach to improve segmentation accuracy and reduce memory usage for storing class probability distribution. We have provided an extensive evaluation on common benchmarks and our own dataset. The results confirm that the developed system performs strongly in terms of sensor pose estimation, surface reconstruction, and segmentation in comparison to other state-of-the-art systems. As future work, we plan on incorporating depth images in the Mask R-CNN pipeline and on reducing the runtime requirements of the proposed system. Devising methods for automatic tuning of the ICP weight in our camera tracking objective function are also promising directions for further investigation.
\bibliographystyle{IEEEtran}
\bibliography{References}

\end{document}